\documentclass[twoside,leqno,twocolumn]{article}

\usepackage[letterpaper]{geometry}

\usepackage{ltexpprt}
\usepackage{balance}
\usepackage{amsmath}
\usepackage[tight,footnotesize]{subfigure}
\usepackage{graphicx}
\usepackage{bm}
\usepackage{algorithm}
\usepackage[noend]{algorithmic}
\usepackage{multirow}
\usepackage{slashbox}
\usepackage{caption}
\usepackage{amsfonts}
\newtheorem{definition}{Definition}
\begin{document}

\title{\Large Adaptive Regularized Submodular Maximization}
\author{Shaojie Tang \thanks{Naveen Jindal School of Management, University of Texas at Dallas}
\and Jing Yuan \thanks{Department of Computer Science, University of Texas at Dallas}}

\date{}

\maketitle

% Copyright Statement
% When submitting your final paper to a SIAM proceedings, it is requested that you include
% the appropriate copyright in the footer of the paper.  The copyright added should be
% consistent with the copyright selected on the copyright form submitted with the paper.
% Please note that "20XX" should be changed to the year of the meeting.

% Default Copyright Statement
\fancyfoot[R]{\scriptsize{Copyright \textcopyright\ 20XX by SIAM\\
Unauthorized reproduction of this article is prohibited}}

% Depending on which copyright you agree to when you sign the copyright form, the copyright
% can be changed to one of the following after commenting out the default copyright statement
% above.

%\fancyfoot[R]{\scriptsize{Copyright \textcopyright\ 20XX\\
%Copyright for this paper is retained by authors}}

%\fancyfoot[R]{\scriptsize{Copyright \textcopyright\ 20XX\\
%Copyright retained by principal author's organization}}

%\pagenumbering{arabic}
%\setcounter{page}{1}%Leave this line commented out.

\begin{abstract} \small\baselineskip=9pt In this paper, we study the problem of  maximizing the difference between an adaptive submodular (revenue) function and an non-negative modular (cost) function under the adaptive setting. The input of our problem is a set of $n$ items, where each item has a particular state drawn from some known prior distribution $p$. The revenue function $g$ is defined over items and states, and the cost function $c$ is defined over items, i.e., each item has a fixed cost. The state of each item is unknown initially, one must select an item in order to observe its realized state. A policy $\pi$ specifies which item to pick next based on the observations made so far. Denote by $g_{avg}(\pi)$ the expected revenue of $\pi$ and  let $c_{avg}(\pi)$ denote the expected cost of $\pi$.  Our objective is to identify the best policy   $\pi^o\in \arg\max_{\pi}g_{avg}(\pi)-c_{avg}(\pi)$ under a $k$-cardinality constraint. Since our objective function can take on both negative and positive values, the existing results of submodular maximization may not be applicable. To overcome this challenge, we develop a series of effective solutions with performance grantees. Let $\pi^o$ denote the optimal policy. For the case when $g$ is adaptive monotone and adaptive submodular, we develop an effective policy $\pi^l$ such that $g_{avg}(\pi^l) - c_{avg}(\pi^l) \geq (1-\frac{1}{e}-\epsilon)g_{avg}(\pi^o) - c_{avg}(\pi^o)$,  using only $O(n\epsilon^{-2}\log \epsilon^{-1})$ value oracle queries. For the case when $g$ is adaptive submodular, we present a randomized policy $\pi^r$ such that  $g_{avg}(\pi^r) - c_{avg}(\pi^r) \geq \frac{1}{e}g_{avg}(\pi^o) - c_{avg}(\pi^o)$.\end{abstract}

\section{Introduction}
Maximizing a submodular function subject to practical constraints has attracted increased attention recently \cite{golovin2011adaptive,tang2020influence,yuan2017no,krause2007near}.  Submodularity  encodes a natural diminishing returns
property, which can be found in a wide variety of machine learning tasks such as active learning \cite{golovin2011adaptive}, virtual marketing \cite{tang2020influence,yuan2017no}, sensor placement \cite{krause2007near}, and data summarization \cite{lin2011class}. Under the non-adaptive setting, where one must select a group of items all at once, \cite{nemhauser1978analysis} show that a classic greedy algorithm achieves $1-1/e$ approximation ratio for the problem of maximizing a monotone and non-negative submodular function subject to a cardinality constraint.  For non-monotone and  non-negative objectives, \cite{buchbinder2014submodular} obtain an approximation  of $1/e+0.004$.

Very recently, \cite{harshaw2019submodular} study the problem of maximizing the difference between a monotone non-negative submodular function and an non-negative modular function. Given that the objective function of the above problem may take both positive and negative values, most existing technologies, which  require the objective function to take only non-negative values,  can not  provide nontrivial approximation guarantees. They overcome this challenge by developing a series of effective algorithms. In this paper, we extend their work to the adaptive setting by considering the  problem of  adaptive regularized submodular maximization, i.e., our goal is to adaptively select a group of items to maximize the difference between an adaptive submodular (revenue) function and an non-negative modular (cost) function. We next provide more details about our adaptive setting. Following the framework of adaptive submodular maximization \cite{golovin2011adaptive}, a natural stochastic variant of the classical non-adaptive submodular maximization problem,  we assume that each item is in a particular state drawn from a known prior distribution $p$. The state of each item is unknown initially, one must select an item before observing its state. A policy $\pi$ specifies which item to pick next based on the observations made so far. Note that the decision on selecting an item is irrevocable, that is, we can not discard any item that is previously selected.  The revenue function $g$ is defined over items and states, and the cost function $c$ is defined over items.  Note that there are two sources of randomness that make our problem more complicated than its non-adaptive counterpart. One is the random realization of items' states, and the other one is the random decision that is made by the policy. We use $g_{avg}(\pi)$ to denote the expected revenue of $\pi$ and  let $c_{avg}(\pi)$ denote the expected cost of $\pi$. Our objective is to identify the best policy:
\begin{eqnarray*}
%\label{pb:1}
\max_{\pi}g_{avg}(\pi)-c_{avg}(\pi)
\end{eqnarray*} under a $k$-cardinality constraint. The above formulation has its applications in many domains. When $g_{avg}(\pi)$ represents the revenue of $\pi$ and $c_{avg}(\pi)$ encodes the cost of $\pi$, the above formulation is to maximize profits. In general,  the above formulation may be interpreted as a regularized submodular maximization problem under the adaptive setting. Since our objective function can take on both negative and positive values, the existing results of submodular maximization, which require the objective function to take only non-negative values, may not be applicable.

%For the case when $g$ is adaptive monotone and adaptive submodular, we develop an effective policy $\pi^l$ such that $g_{avg}(\pi^l) - c_{avg}(\pi^l) \geq (1-\frac{1}{e}-\epsilon)g_{avg}(\pi^o) - c_{avg}(\pi^o)$,  using only $O(n\epsilon^{-2}\log \epsilon^{-1})$ value oracle queries. For the case when $g$ is adaptive submodular, we present a randomized policy $\pi^r$ such that  $g_{avg}(\pi^r) - c_{avg}(\pi^r) \geq \frac{1}{e}g_{avg}(\pi^o) - c_{avg}(\pi^o)$. %Their goal is to adaptively select a group of $k$ items so as to maximize the expected revenue of an adaptive submodular and adaptive monotone function. There have been numerous research studies on adaptive submodular maximization under different settings \cite{chen2013near,tang2020influence,tang2020price,yuan2017adaptive,fujii2019beyond}.
Our contribution is threefold. We first consider the case when the revenue function   $g$ is  adaptive monotone and adaptive submodular. We develop an effective policy $\pi^d$ such that $g_{avg}(\pi^d) - c_{avg}(\pi^d) \geq (1-\frac{1}{e})g_{avg}(\pi^o) - c_{avg}(\pi^o)$,  using $O(kn)$ value oracle queries. Our second result is the development of a faster policy $\pi^l$ such that $g_{avg}(\pi^l) - c_{avg}(\pi^l) \geq (1-\frac{1}{e}-\epsilon)g_{avg}(\pi^o) - c_{avg}(\pi^o)$,  using only $O(n\epsilon^{-2}\log \epsilon^{-1})$ value oracle queries. For the case when $g$ is adaptive submodular, we present a randomized policy $\pi^r$ such that  $g_{avg}(\pi^r) - c_{avg}(\pi^r) \geq \frac{1}{e}g_{avg}(\pi^o) - c_{avg}(\pi^o)$.

\section{Related Work}
Submodular maximization is a well-studied topic due to its applications in a wide range of domains including active learning \cite{golovin2011adaptive}, virtual marketing \cite{tang2020influence,yuan2017no}, sensor placement \cite{krause2007near}. Most of existing studies focus on the non-adaptive setting where one must select a group of items all at once. \cite{nemhauser1978analysis} show that a classic greedy algorithm, which iteratively selects the item that has the largest marginal revenue  on top of the previously selected items, achieves a $1-1/e$ approximation ratio when maximizing a monotone non-negative submodular function subject to a cardinality constraint. The problem of maximizing a sum of a non-negative monotone submodular function and  an (arbitrary) modular function is first studied in \cite{sviridenko2017optimal}. Notably, their objective function  may take on negative values. \cite{feldman2020guess} develop a faster algorithm using a surrogate objective that varies with time. For the case of a cardinality constraint and a non-negative $c$, \cite{harshaw2019submodular} develop the first practical algorithm. Recently, \cite{kazemi2020regularized} extend this study to streaming and distributed settings. Our work is different from theirs in that we focus on  the adaptive setting. Moreover, we consider a more general problem of maximizing the difference of an non-negative \emph{non-monotone} adaptive submodular function and  an non-negative modular function.

\section{Preliminaries}
%We start by introducing some important notations. In the rest of this paper, we use  $[m]$ to denote the set $\{1, 2, \cdots, m\}$, and we use $|X|$ to denote the cardinality of a set $X$.
In the rest of this paper, we use  $[m]$ to denote the set $\{0, 1, \cdots, m\}$.

\subsection{Items and States} The input of our problem is a set  $E$ of $n$ items, where each item $e \in E$ is in a particular state from  $O$.  We use a function $\phi: E\rightarrow O$, called a \emph{realization}, to represent the states of all items, i.e., $\phi(i)$ represents the realization of $e$'s state. Denote by $\Phi=\{\Phi(e) \mid e\in E\}$  the  random realizations of $E$, where $\Phi(e) \in O$ represents a random realization of $e$. There is a known prior probability distribution $p=\{\Pr[\Phi=\phi]: \phi\in U\}$ over realizations $U$.  The state $\Phi(e)$ of each item $e \in E$ is  unknown initially, one must select $e$ before observing its realized state. If we select multiple items, we are able to observe a \emph{partial realization} of their states.   Given any partial realization $\psi$, we define the \emph{domain} $\mathrm{dom}(\psi)$ of $\psi$  the group of all items involved in $\psi$, i.e., $\psi = \cup_{e\in  \mathrm{dom}(\psi)} \{\phi(e)\}$. A partial realization $\psi$ is said to be consistent with a realization $\phi$, denoted $\phi \sim \psi$, if they are equal everywhere in $\mathrm{dom}(\psi)$.  A partial realization $\psi$  is said to be a \emph{subrealization} of  $\psi'$, denoted  $\psi \subseteq \psi'$, if $\mathrm{dom}(\psi) \subseteq \mathrm{dom}(\psi')$ and they are equal everywhere in the domain $\mathrm{dom}(\psi)$ of $\psi$. Let $p(\phi\mid \psi)$ denote the conditional distribution over realizations conditioned on  a partial realization $\psi$: $p(\phi\mid \psi) =\Pr[\Phi=\phi\mid \Phi\sim \psi ]$.

\subsection{Revenue and Cost}For a set $Y\subseteq E$ of items and a realization $\phi$, let $g(Y, \phi)$ represent the revenue of selecting $Y$ conditioned on $\phi$, where $g$ is called \emph{revenue function}.  Moreover, each item $e\in E$ has a fixed cost $c_e$. For any set $Y\subseteq E$ of items, let $c(Y) = \sum_{e\in Y} c_e$ denote the total cost of $Y$, where $c$ is called \emph{cost function}.

%After selecting a set of items, we are able to observe a \emph{partial realization} of those items' states.   For any partial realization $\psi$, we define the \emph{domain} $\mathrm{dom}(\psi)$ of $\psi$  the set of all items involved in $\psi$. We say a partial realization $\psi$ is consistent with a realization $\phi$, denoted $\phi \sim \psi$, if they are equal everywhere in $\mathrm{dom}(\psi)$. Moreover, we say $\psi$  is a \emph{subrealization} of  $\psi'$, denoted  $\psi \subseteq \psi'$, if $\mathrm{dom}(\psi) \subseteq \mathrm{dom}(\psi')$ and they are equal everywhere in the domain $\mathrm{dom}(\psi)$ of $\psi$. Let $p(\phi\mid \psi)$ denote the conditional distribution over realizations conditioned on  a partial realization $\psi$: $p(\phi\mid \psi) =\Pr[\Phi=\phi\mid \Phi\sim \psi ]$.

\subsection{Problem Formulation} %We consider an adaptive optimization problem where we sequentially select a group of items, after each selection, we observe the partial realization of the states of  those items which have been previously selected.
A policy  specifies  which item to select next based on  the partial realization observed  so far. Mathematically, we represent a policy using a function $\pi$ that maps a set of observations  to a distribution $\mathcal{P}(E)$ of $E$: $\pi: 2^{E}\times O^E \rightarrow \mathcal{P}(E)$.

\begin{definition}[Policy  Concatenation]
Given two policies $\pi$ and $\pi'$,  let $\pi @\pi'$ denote a policy that runs $\pi$ first, and then runs $\pi'$, ignoring the observation obtained from running $\pi$.
\end{definition}

\begin{definition}[Level-$t$-Truncation of a Policy]
Given a policy $\pi$, we define its  level-$t$-truncation $\pi_t$  as a policy that runs $\pi$ until it selects $t$ items.
\end{definition}

 For each  realization $\phi$, let $E(\pi, \phi)$ denote the subset of items selected by $\pi$ under realization $\phi$. Note that $E(\pi, \phi)$ is a random variable. The expected  revenue $g_{avg}(\pi)$ of a policy $\pi$ can be written as
\[g_{avg}(\pi)=\mathbb{E}_{\Phi\sim p, \Pi}[g(E(\pi, \Phi), \Phi)]\]
 where the expectation is taken over possible realizations and random outputs of the policy.  Similarly,  the expected  cost $c_{avg}(\pi)$ of a policy $\pi$ can be written as
\[c_{avg}(\pi)=\mathbb{E}_{\Phi\sim p, \Pi}[c(E(\pi, \Phi))]\]

We next introduce the the conditional expected marginal revenue $g(e \mid \psi)$ of $e$ conditioned on a partial realization $\psi$:
 \[g(e \mid \psi)=\mathbb{E}_{\Phi}[g(\mathrm{dom}(\psi)\cup \{e\}, \Phi)-g(\mathrm{dom}(\psi), \Phi)\mid \Phi \sim \psi]\] where the expectation is taken over $\Phi$ with respect to $p(\phi\mid \psi)=\Pr(\Phi=\phi \mid \Phi \sim \psi)$
\begin{definition}[Adaptive Submodularity] For any two partial realizations $\psi$ and $\psi'$ such that $\psi\subseteq \psi'$, we assume that the following holds for each $e\in E\setminus \mathrm{dom}(\psi')$:
\begin{eqnarray}\label{def:33}
g(e\mid \psi) \geq g(e\mid \psi')
\end{eqnarray}
\end{definition}

Let $\Omega$  denote the set of all policies that select at most $k$ items, our objective is listed in below:
\begin{eqnarray}
\label{pb:1}
\max_{\pi\in \Omega} g_{avg}(\pi)-c_{avg}(\pi)
\end{eqnarray}

 Before presenting our solutions to the above problem, we introduce some additional notations. By abuse of notation,   for any partial realization $\psi$, we define $g(\psi)=\mathbb{E}_{\Phi\sim \psi} [g(\mathrm{dom}(\psi), \Phi)]$.
 We next introduce two useful functions: $G_i$, the distorted objective function,  and $H_i$, which is used to analyze
the trajectory of $G_i$. For any partial realization $\psi$, and any iteration $i\in [k]$ of our algorithms, we define
\[G_i(\psi) = (1-\frac{1}{k})^{k-i} g(\psi)-c(\textrm{dom}(\psi))\]
For any partial realization $\psi$, and any iteration $i\in [k-1]$ of our algorithms, we define
\[H_i(\psi, e) =(1-\frac{1}{k})^{k-(i+1)} g(e \mid \psi)-c_e\]

\section{Monotone $g$: Adaptive Distorted Greedy Policy}
We start with the case when $g$ is adaptive submodular and \emph{adaptive monotone} \cite{golovin2011adaptive}, i.e.,  for any realization $\psi$, the following holds for each $e\in E\setminus \mathrm{dom}(\psi)$: $g(e\mid \psi) \geq 0$.  Our approach is a natural extension of the \emph{Distroted-Greedy} algorithm, the first practical non-adaptive algorithm developed by \cite{harshaw2019submodular}. Note that there are two factors that make our problem more complicated than its non-adaptive counterpart. First, since the objective function is defined over random realization, the key of analysis is to estimate the expected utility under the distribution of realizations $p$. Second, the policy itself might produce random outputs even under the same realization, this adds an additional layer of difficulty to the design and analysis of our policy. To address the above complications, We develop an \emph{Adaptive Distorted Greedy Policy} $\pi^d$ such that $g_{avg}(\pi^d) - c_{avg}(\pi^d) \geq (1-\frac{1}{e})g_{avg}(\pi^o) - c_{avg}(\pi^o)$, where $\pi^o$ denotes the optimal policy. We next explain the idea of $\pi^d$ (Algorithm \ref{alg:LPP1}), then analyze its performance bound.

\subsection{Design of $\pi^d$} We first add a dummy item $d$ to the ground set, such that, $c_d=0$, and for any partial realization $\psi$, we have $g(d \mid \psi) =0$. Let $E'=E\cup \{d\}$. We add this to ensure that our policy will not select an item that has an negative profit. Note that $d$ can be safely removed from the final solution with affecting its performance. $\pi^{d}$  performs in $k$ iterations: It starts with an empty set. In each iteration $i\in [k-1]$, let $\psi_i$ denote the current partial realization, $\pi^{d}$ selects an item $e_i$ that maximizes $H_i(\psi_i, \cdot)$:
       \[e_i\leftarrow \arg\max_{e \in E'}H_i(\psi_i, e)\]
After observing the state $\phi(e_i)$ of $e_i$, update  the current partial realization $\psi_{i+1}$ using  $\psi_{i}\cup\{\phi(e_i)\}$.  This process iterates until all  $k$ items have been selected.

\begin{algorithm}[hptb]
\caption{Adaptive Distorted Greedy Policy $\pi^d$}
\label{alg:LPP1}
\begin{algorithmic}[1]
\STATE $S_0=\emptyset; i=0; \psi_0=\emptyset$.
\WHILE {$i < k$}
%\STATE observe $\psi_{i}$;
\STATE $e_i\leftarrow \arg\max_{e \in E'}H_i(\psi_i, e)$;
\STATE $S_i\leftarrow S_{i-1}\cup \{e_i\}$;
\STATE $\psi_{i+1}\leftarrow \psi_{i}\cup\{\phi(e_i)\}$;  $i\leftarrow i+1$;
\ENDWHILE
\RETURN $S_k$
\end{algorithmic}
\end{algorithm}

\subsection{Performance Analysis}
We first present three preparatory lemmas which are used to lower bound the marginal gain in the distorted objective. 
\begin{lemma}
\label{lem:a1}
In each iteration of $\pi^d$,
\begin{eqnarray*}
&&\mathbb{E}_{\Phi\sim \psi_i}[G_{i+1}(\Psi_{i+1})- G_{i}(\psi_i)] \\
&&= H_i(\psi_i, e_i)+\frac{1}{k}(1-\frac{1}{k})^{k-(i+1)} g(\psi_i)
\end{eqnarray*}
\end{lemma}
\emph{Proof:} We start with the case when $e_i\in \textrm{dom}(\psi_i)$,
\begin{eqnarray*}
&& \mathbb{E}_{\Phi\sim \psi_i}[G_{i+1}(\Psi_{i+1})- G_{i}(\psi_i)] \\
&&= (1-\frac{1}{k})^{k-(i+1)} g(\psi_i)-(1-\frac{1}{k})^{k-i} g(\psi_i)\\
&& = (1-\frac{1}{k})^{k-(i+1)} g(\psi_i)\\
&&\quad\quad-(1-\frac{1}{k}) (1-\frac{1}{k})^{k-(i+1)} g(\psi_i)\\
&& = \frac{1}{k}(1-\frac{1}{k})^{k-(i+1)} g(\psi_i) \\
&&= H_i(\psi_i, e_i)+\frac{1}{k}(1-\frac{1}{k})^{k-(i+1)} g(\psi_i)
\end{eqnarray*}
The last equality is due to $ H_i(\psi_i, e_i) = 0$ when $e_i\in \textrm{dom}(\psi_i)$. We next prove the case  when $e_i\notin \textrm{dom}(\psi_i)$,
\begin{eqnarray*}
&&\mathbb{E}_{\Phi\sim \psi_i}[G_{i+1}(\Psi_{i+1})- G_{i}(\psi_i)]\\
&& = \mathbb{E}_{\Phi\sim \psi_i}[(1-\frac{1}{k})^{k-(i+1)} g(\psi_i\cup \{\Phi(e_i)\})\\
&&\quad\quad-c(\textrm{dom}(\psi_i)\cup\{e_i\}) \\
&&\quad\quad- ((1-\frac{1}{k})^{k-i} g(\psi_i)-c(\textrm{dom}(\psi_i)))]\\
&& = \mathbb{E}_{\Phi\sim \psi_i}[(1-\frac{1}{k})^{k-(i+1)} g(\psi_i\cup \{\Phi(e_i)\})]\\
&&\quad\quad-c(\textrm{dom}(\psi_i)\cup\{e_i\})\\
&&\quad\quad- ((1-\frac{1}{k})^{k-i} g(\psi_i)-c(\textrm{dom}(\psi_i)))\\
&& = \mathbb{E}_{\Phi\sim \psi_i}[(1-\frac{1}{k})^{k-(i+1)} g(\psi_i\cup \{\Phi(e_i)\})]\\
&&\quad\quad -c(\textrm{dom}(\psi_i) \cup\{e_i\})\\
&&\quad\quad- ((1-\frac{1}{k})^{k-(i+1)} (1-\frac{1}{k}) g(\psi_i)-c(\textrm{dom}(\psi_i))\\
&& =  \mathbb{E}_{\Phi\sim \psi_i}[(1-\frac{1}{k})^{k-(i+1)} (g(\psi_i\cup \{\Phi(e_i)\}) -  g(\psi_i))]\\
 &&\quad\quad - c_{e_i} + \frac{1}{k}(1-\frac{1}{k})^{k-(i+1)}g(\psi_i)\\
 && = g(e_i\mid \psi_i) - c_{e_i} + \frac{1}{k}(1-\frac{1}{k})^{k-(i+1)}g(\psi_i)\\
&& = H_i(\psi_i, e_i)+\frac{1}{k}(1-\frac{1}{k})^{k-(i+1)} g(\psi_i)
\end{eqnarray*} The fourth equality is due to $e_i\notin \textrm{dom}(\psi_i)$. $\Box$

\begin{lemma}
\label{lem:a2}
In each iteration of $\pi^d$,
\begin{eqnarray*}
H_i(\psi_i, e_i)&\geq \frac{1}{k}(1-\frac{1}{k})^{k-(i+1)} \mathbb{E}_{\Phi\sim \psi_i}[g_{avg}(\pi^o)-g_{avg}(\pi^d_i)]\\
&-\frac{1}{k}\mathbb{E}_{\Phi\sim \psi_i}[c_{avg}(\pi^o)]
\end{eqnarray*}
\end{lemma}
\emph{Proof:} Let $A_e$ be in indicator that $e$ is selected by the optimal solution $\pi^o$ conditioned on a partial realization $\psi_i$, then we have
\begin{eqnarray*}
&& H_i(\psi_i, e_i) = (1-\frac{1}{k})^{k-(i+1)} g(e_i \mid \psi_i)-c_{e_i}\\
&& = \max_{e\in E'}[(1-\frac{1}{k})^{k-(i+1)} g(e \mid \psi_i)-c_{e}]\\
&& \geq \frac{1}{k}\sum_{e\in E'}\Pr[A_e=1][(1-\frac{1}{k})^{k-(i+1)} g(e \mid \psi_i)-c_{e}]\\
&& \geq \frac{1}{k}\sum_{e\in E'}[(1-\frac{1}{k})^{k-(i+1)} (g_{avg}(\pi^o)-g_{avg}(\pi^d_i))\\
&&\quad\quad-c_{avg}(\pi^o)]\\
&& = \frac{1}{k}(1-\frac{1}{k})^{k-(i+1)} \mathbb{E}_{\Phi\sim \psi_i}[g_{avg}(\pi^o)-g_{avg}(\pi^d_i)]\\
&&\quad\quad-\frac{1}{k}\mathbb{E}_{\Phi\sim \psi_i}[c_{avg}(\pi^o)]
\end{eqnarray*} The second equality is due to the design of $\pi^d$, i.e., it selects an item $e_i$ that maximizes $H_i(\psi_i, \cdot)$. The first inequality is due to $\sum_{e\in E'}\Pr[A_e=1]\leq k$ since $\pi^o$ selects at most $k$ items. The second inequality is due to $g$ is adaptive submodular and $c$ is modular. $\Box$

\begin{lemma}
\label{lem:a3}
In each iteration of $\pi^d$,
\begin{eqnarray*}
&&\mathbb{E}_{\Psi_i}[\mathbb{E}_{\Phi\sim \Psi_i}[G_{i+1}(\Psi_{i+1})- G_{i}(\Psi_i)]] \\
&&\geq \frac{1}{k}(1-\frac{1}{k})^{k-(i+1)}g_{avg}(\pi^o) -\frac{1}{k}c_{avg}(\pi^o)
\end{eqnarray*}
\end{lemma}

\emph{Proof:} We first prove that for any partial realization $\psi_i$, the following inequality holds:
\begin{eqnarray}
&\mathbb{E}_{\Phi\sim \psi_i}[G_{i+1}(\Psi_{i+1})- G_{i}(\psi_i)]\label{eq:a1}\\
&\geq \frac{1}{k}(1-\frac{1}{k})^{k-(i+1)} \mathbb{E}_{\Phi\sim \psi_i}[g_{avg}(\pi^o)]-\frac{1}{k}\mathbb{E}_{\Phi\sim \psi_i}[c_{avg}(\pi^o)]~\nonumber
\end{eqnarray}
Due to Lemma \ref{lem:a1}, we have
\begin{eqnarray*}
&&\mathbb{E}_{\Phi\sim \psi_i}[G_{i+1}(\Psi_{i+1})- G_{i}(\psi_i)] \\
&&= H_i(\psi_i, e_i)+\frac{1}{k}(1-\frac{1}{k})^{k-(i+1)} g(\psi_i)\\
&& \geq \frac{1}{k}(1-\frac{1}{k})^{k-(i+1)} \mathbb{E}_{\Phi\sim \psi_i}[g_{avg}(\pi^o)-g_{avg}(\pi^d_i)]\\
&&\quad\quad-\frac{1}{k}\mathbb{E}_{\Phi\sim \psi_i}[c_{avg}(\pi^o)] + \frac{1}{k}(1-\frac{1}{k})^{k-(i+1)} g(\psi_i)\\
&& = \frac{1}{k}(1-\frac{1}{k})^{k-(i+1)} \mathbb{E}_{\Phi\sim \psi_i}[g_{avg}(\pi^o)]\\
&&\quad\quad-\frac{1}{k}\mathbb{E}_{\Phi\sim \psi_i}[c_{avg}(\pi^o)]
\end{eqnarray*}
The inequality is due to Lemma \ref{lem:a2} and the second equality is due to $\mathbb{E}_{\Phi\sim \psi_i}[g_{avg}(\pi^d_i)] =  g(\psi_i)$. Now we are ready to prove this lemma.
\begin{eqnarray*}
&&\mathbb{E}_{\Psi_i}[\mathbb{E}_{\Phi\sim \Psi_i}[G_{i+1}(\Psi_{i+1})- G_{i}(\Psi_i)]] \\
&& \geq \mathbb{E}_{\Psi_i}[\frac{1}{k}(1-\frac{1}{k})^{k-(i+1)} \mathbb{E}_{\Phi\sim \Psi_i}[g_{avg}(\pi^o)]\\
&&\quad\quad-\frac{1}{k}\mathbb{E}_{\Phi\sim \Psi_i}[c_{avg}(\pi^o)]]\\
&& = \mathbb{E}_{\Psi_i}[\frac{1}{k}(1-\frac{1}{k})^{k-(i+1)} \mathbb{E}_{\Phi\sim \Psi_i}[g_{avg}(\pi^o)]]\\
&& \quad\quad-  \mathbb{E}_{\Psi_i}[\frac{1}{k}\mathbb{E}_{\Phi\sim \Psi_i}[c_{avg}(\pi^o)]]\\
&& = \frac{1}{k}(1-\frac{1}{k})^{k-(i+1)}g_{avg}(\pi^o) -\frac{1}{k}c_{avg}(\pi^o)
\end{eqnarray*}
The  inequality is due to (\ref{eq:a1}). $\Box$

We next present the first main theorem of this paper.
\begin{theorem}
$g_{avg}(\pi^d) - c_{avg}(\pi^d) \geq (1-\frac{1}{e})g_{avg}(\pi^o) - c_{avg}(\pi^o)$.
\end{theorem}
\emph{Proof:} According to the definition of $G_k$, we have $\mathbb{E}_{\Psi_k}[G_{k}(\Psi_k)] = \mathbb{E}_{\Psi_k}[(1-\frac{1}{k})^{0} g(\Psi_k)-c(\textrm{dom}(\Psi_k))]=g_{avg}(\pi^d) - c_{avg}(\pi^d)$ and $\mathbb{E}_{\Psi_0}[G_{0}(\Psi_0)] = \mathbb{E}_{\Psi_0}[(1-\frac{1}{k})^{k} g(\Phi(S_0))-c(\textrm{dom}(\Psi_0))]=0$. Hence,
\begin{eqnarray*}
&& g_{avg}(\pi^d) - c_{avg}(\pi^d) = \mathbb{E}_{\Psi_k}[G_{k}(\Psi_k)] - \mathbb{E}_{\Psi_0}[G_{0}(\Psi_0)]\\
&& = \mathbb{E}_{\Psi_{k-1}}[\mathbb{E}_{\Phi \sim \Psi_{k-1}}[G_{k}(\Psi_k)]]- \mathbb{E}_{\Psi_0}[\mathbb{E}_{\Phi\sim \Psi_0}[G_{0}(\Psi_0)]] \\
&& = \sum_{i\in [k-1]}( \mathbb{E}_{\Psi_i}[\mathbb{E}_{\Phi\sim \Psi_i}[G_{i+1}(\Psi_{i+1})]]\\
&&\quad\quad- \mathbb{E}_{\Psi_i}[\mathbb{E}_{\Phi\sim \Psi_i}[G_{i}(\Psi_i)]]) \\
&& = \sum_{i\in [k-1]} \mathbb{E}_{\Psi_i}[\mathbb{E}_{\Phi\sim \Psi_i}[G_{i+1}(\Psi_{i+1})- G_{i}(\Psi_i)]] \\
&& \geq \sum_{i\in [k-1]} (\frac{1}{k}(1-\frac{1}{k})^{k-(i+1)}g_{avg}(\pi^o) -\frac{1}{k}c_{avg}(\pi^o))\\
&& = \sum_{i\in [k-1]}(\frac{1}{k}(1-\frac{1}{k})^{k-(i+1)}g_{avg}(\pi^o)) - c_{avg}(\pi^o)\\
&& \geq (1-\frac{1}{e})g_{avg}(\pi^o) - c_{avg}(\pi^o)
\end{eqnarray*} The first inequality is due to Lemma \ref{lem:a3}.
$\Box$
\section{Monotone $g$: Linear-time Adaptive Distorted Greedy Policy}
We next propose a faster algorithm \emph{Linear-time Adaptive Distorted Greedy Policy}, denoted by  $\pi^{l}$, for the case when $g$ is adaptive monotone. As compared with $\pi^{d}$ whose running time is $O(nk)$, our new policy $\pi^{l}$ achieves nearly the same performance guarantee with $O(n \log  \frac{1}{\epsilon})$ value oracle queries. Our design is inspired by the sampling technique developed in \cite{mirzasoleiman2015lazier} for maximizing a monotone and submodular function. Very recently, \cite{tang2021beyond} extends this approach to the adaptive setting to develop a linear-time adaptive policy for maximizing an adaptive submodular and adaptive monotone function. In this work, we apply this technique to design a linear-time adaptive policy for our adaptive regularized submodular maximization problem. Note that our objectives are not adaptive monotone and they may take negative values. %We next present our adaptive policy \emph{Adaptive Stochastic Greedy}, denoted by  $\pi^{asg}$.

\subsection{Design of $\pi^l$} We present the details of our algorithm in Algorithm \ref{alg:LPP13}. We first add a set $D$ of $k-1$ dummy items to the ground set, such that, each dummy item $d\in D$ has zero cost, i.e., $\forall d\in D, c_d=0$, and for any $d \in D$, and any
partial realization $\psi$, we have $g(d \mid \psi) =0$. Let $E'=E\cup D$. We next explain the idea of $\pi^{l}$: It starts with an empty set. In each iteration $i\in[k-1]$,  $\pi^l$ first samples a set $R_i$ of size $\frac{n}{k}\log\frac{1}{\epsilon}$ uniformly at random,  then adds an item $e_i$ with the largest $H_i(\psi_i, \cdot)$ from $R_i$ to the solution. After observing the state $\phi(e_i)$ of $e_i$, update  the current partial realization $\psi_{i+1}$ using  $\psi_{i}\cup\{\phi(e_i)\}$.  This process iterates until all  $k$ items have been selected. %Our approach is a natural extension of the \emph{Stochastic Greedy} algorithm \cite{mirzasoleiman2015lazier}, the first linear-time algorithm for maximizing a monotone submodular function under the non-adaptive setting, we generalize their results to the adaptive setting.
%We will show that $\pi^{d}$  achieves $1-1/e-\epsilon$ approximation ratio for maximizing an adaptive monotone and submodular function, and it has linear running %time independent of the cardinality constraint $k$.
It was worth noting that the technique of lazy updates \cite{minoux1978accelerated}  can be used to further accelerate the computation of our algorithms in practice.

\begin{algorithm}[hptb]
\caption{Linear-time Adaptive Distorted Greedy Policy $\pi^{l}$}
\label{alg:LPP13}
\begin{algorithmic}[1]
\STATE $S_0=\emptyset; i=0; \psi_0=\emptyset$.
\WHILE {$i < k$}
%\STATE observe $\psi_i$;
\STATE $R_i\leftarrow$ a random set sampled uniformly at random  from $E'$;
\STATE $e_i\leftarrow \arg\max_{e \in R_i}H_i(\psi_i, e)$;
\STATE $S_i\leftarrow S_{i-1}\cup \{e_i\}$;
\STATE $\psi_{i+1}\leftarrow \psi_{i}\cup\{\phi(e_i)\}$;  $i\leftarrow i+1$;
\ENDWHILE
\RETURN $S_k$
\end{algorithmic}
\end{algorithm}

\subsection{Performance Analysis}
We first present three preparatory lemmas.
\begin{lemma}
\label{lem:b1}
In each iteration of $\pi^d$,
\begin{eqnarray*}
&&\mathbb{E}_{\Phi\sim \psi_i}[G_{i+1}(\Psi_{i+1})- G_{i}(\psi_i)] \\
&&= \mathbb{E}_{e_i}[H_i(\psi_i, e_i)]+\frac{1}{k}(1-\frac{1}{k})^{k-(i+1)} g(\psi_i)
\end{eqnarray*}
\end{lemma}
The above lemma immediately follows from Lemma \ref{lem:a1}.

\begin{lemma}
\label{lem:b2}
In each iteration of $\pi^d$,
$
\mathbb{E}_{e_i}[H_i(\psi_i, e_i)] \geq \frac{1}{k}(1-\frac{1}{k})^{k-(i+1)} \mathbb{E}_{\Phi\sim \psi_i}[g_{avg}(\pi^o)-g_{avg}(\pi^d_i)]-\frac{1}{k}\mathbb{E}_{\Phi\sim \psi_i}[c_{avg}(\pi^o)]$.
\end{lemma}
\emph{Proof:} Let $B_e$ be an indicator that $e$ is selected by $\pi^l$ in iteration $i$ conditioned on partial realization $\psi_i$. Let $C_e$ be an indicator that $R_i \cap  M(\psi_i) \neq \emptyset$, where $M(\psi_i)$ contains the $k$ items with the largest marginal contribution to $\psi_i$ in terms of $H_i(\psi_i, \cdot)$, i.e., $M(\psi_i) \leftarrow \arg\max_{S\subseteq E', |S|\leq k} \sum_{e\in S} H_i(\psi_i, e)$. Then we have
\begin{eqnarray*}
&& \mathbb{E}_{e_i}[H_i(\psi_i, e_i)]  \\
&& = \sum_{e\in E'} \Pr[B_e=1] ((1-\frac{1}{k})^{k-(i+1)} g(e \mid \psi_i)-c_{e}) \\
&& \geq \Pr[C_e=1] \frac{1}{k}\sum_{e\in M(\psi_i)} ((1-\frac{1}{k})^{k-(i+1)} g(e \mid \psi_i)-c_{e})\\
&& \geq (1-\epsilon) \frac{1}{k}\sum_{e\in M(\psi_i)} ((1-\frac{1}{k})^{k-(i+1)} g(e \mid \psi_i)-c_{e})\\
&& \geq (1-\epsilon) \frac{1}{k}\sum_{e\in E'} \Pr[A_e=1] ((1-\frac{1}{k})^{k-(i+1)} g(e \mid \psi_i)-c_{e})\\
&& \geq (1-\epsilon)\frac{1}{k}(1-\frac{1}{k})^{k-(i+1)}\mathbb{E}_{\Phi\sim \psi_i}[g_{avg}(\pi^o)-g_{avg}(\pi^d_i)]\\
&& \quad\quad-(1-\epsilon)\frac{1}{k}\mathbb{E}_{\Phi\sim \psi_i}[c_{avg}(\pi^o)]
\end{eqnarray*} The second inequality is due to Lemma 4 in \cite{tang2021beyond}, where they show that $\Pr[R_i \cap  M(\psi_i) \neq \emptyset]\geq 1-\epsilon$ given that $R_i$ has size of $\frac{n}{k}\log\frac{1}{\epsilon}$. The third inequality is due to  $\sum_{e\in E'}\Pr[A_e=1]\leq k$. The last inequality is due to $g$ is adaptive submodular and $c$ is modular. $\Box$

\begin{lemma}
\label{lem:b3}
In each iteration of $\pi^d$,
$
\mathbb{E}_{\Psi_i}[\mathbb{E}_{\Phi\sim \Psi_i}[G_{i+1}(\Psi_{i+1})- G_{i}(\Psi_i)]]\geq (1-\epsilon)\frac{1}{k}(1-\frac{1}{k})^{k-(i+1)}g_{avg}(\pi^o) -(1-\epsilon)\frac{1}{k}c_{avg}(\pi^o)$.
\end{lemma}
\emph{Proof:} We first show that for any partial realization $\psi_i$,
\begin{eqnarray}
&\mathbb{E}_{\Phi\sim \psi_i}[G_{i+1}(\Psi_{i+1})- G_{i}(\psi_i)] \label{eq:b1}\\
&\geq (1-\epsilon)(\frac{1}{k}(1-\frac{1}{k})^{k-(i+1)} \mathbb{E}_{\Phi\sim \psi_i}[g_{avg}(\pi^o)]~\nonumber\\
&-\frac{1}{k}\mathbb{E}_{\Phi\sim \psi_i}[c_{avg}(\pi^o)])~\nonumber
\end{eqnarray}
 Due to Lemma \ref{lem:b1}, we have
\begin{eqnarray*}
&&\mathbb{E}_{\Phi\sim \psi_i}[G_{i+1}(\Psi_{i+1})- G_{i}(\psi_i)] \\
&& = \mathbb{E}_{e_i}[H_i(\psi_i, e_i)]+\frac{1}{k}(1-\frac{1}{k})^{k-(i+1)} g(\psi_i)\\
&& \geq (1-\epsilon)\frac{1}{k}(1-\frac{1}{k})^{k-(i+1)}\mathbb{E}_{\Phi\sim \psi_i}[g_{avg}(\pi^o)-g_{avg}(\pi^d_i)]\\
&&\quad\quad -(1-\epsilon)\frac{1}{k}\mathbb{E}_{\Phi\sim \psi_i}[c_{avg}(\pi^o)]+\frac{1}{k}(1-\frac{1}{k})^{k-(i+1)} g(\psi_i)\\
&& = (1-\epsilon)(\frac{1}{k}(1-\frac{1}{k})^{k-(i+1)} \mathbb{E}_{\Phi\sim \psi_i}[g_{avg}(\pi^o)]\\
&&\quad\quad-\frac{1}{k}\mathbb{E}_{\Phi\sim \psi_i}[c_{avg}(\pi^o)] ) +\epsilon\frac{1}{k}(1-\frac{1}{k})^{k-(i+1)} g(\psi_i)\\
&& \geq (1-\epsilon)(\frac{1}{k}(1-\frac{1}{k})^{k-(i+1)} \mathbb{E}_{\Phi\sim \psi_i}[g_{avg}(\pi^o)]\\
&&\quad\quad -\frac{1}{k}\mathbb{E}_{\Phi\sim \psi_i}[c_{avg}(\pi^o)])
\end{eqnarray*}
The first inequality is due to Lemma \ref{lem:b2}, the second equality is due to $\mathbb{E}_{\Phi\sim \psi_i}[g_{avg}(\pi^d_i)] =  g(\psi_i)$, and the last inequality is due to $g$ is non-negative. Now we are ready to prove this lemma.
\begin{eqnarray*}
&&\mathbb{E}_{\Psi_i}[\mathbb{E}_{\Phi\sim \Psi_i}[G_{i+1}(\Psi_{i+1})- G_{i}(\Psi_i)]] \\
&& \geq (1-\epsilon)\mathbb{E}_{\Psi_i}[\frac{1}{k}(1-\frac{1}{k})^{k-(i+1)} \mathbb{E}_{\Phi\sim \Psi_i}[g_{avg}(\pi^o)]\\
&&\quad\quad-\frac{1}{k}\mathbb{E}_{\Phi\sim \Psi_i}[c_{avg}(\pi^o)]]\\
&& = (1-\epsilon)\mathbb{E}_{\Psi_i}[\frac{1}{k}(1-\frac{1}{k})^{k-(i+1)} \mathbb{E}_{\Phi\sim \Psi_i}[g_{avg}(\pi^o)]] \\
&&\quad\quad-  \mathbb{E}_{\Psi_i}[\frac{1}{k}\mathbb{E}_{\Phi\sim \Psi_i}[c_{avg}(\pi^o)]]\\
&& = (1-\epsilon)\frac{1}{k}(1-\frac{1}{k})^{k-(i+1)}g_{avg}(\pi^o) \\
&&\quad\quad-(1-\epsilon)\frac{1}{k}c_{avg}(\pi^o)
\end{eqnarray*}
The first inequality is due to (\ref{eq:b1}). $\Box$

We next present the second main theorem of this paper.
\begin{theorem}
$g_{avg}(\pi^d) - c_{avg}(\pi^d) \geq (1-\frac{1}{e}-\epsilon)g_{avg}(\pi^o) - c_{avg}(\pi^o)$.
\end{theorem}
\emph{Proof:} According to the definition of $G_k$, we have $\mathbb{E}_{\Psi_k}[G_{k}(\Psi_k)] = \mathbb{E}_{\Psi_k}[(1-\frac{1}{k})^{0} g(\Psi_k)-c(\textrm{dom}(\Psi_k))]=g_{avg}(\pi^d) - c_{avg}(\pi^d)$ and $\mathbb{E}_{\Psi_0}[G_{0}(\Psi_0)] = \mathbb{E}_{\Psi_0}[(1-\frac{1}{k})^{k} g(\Phi(S_0))-c(\textrm{dom}(\Psi_0))]=0$. Hence,
\begin{eqnarray*}
&& g_{avg}(\pi^d) - c_{avg}(\pi^d) \\
&& = \mathbb{E}_{\Psi_k}[G_{k}(\Psi_k)] - \mathbb{E}_{\Psi_0}[G_{0}(\Psi_0)]\\
&& = \mathbb{E}_{\Psi_{k-1}}[\mathbb{E}_{\Phi \sim \Psi_{k-1}}[G_{k}(\Psi_k)]]- \mathbb{E}_{\Psi_0}[\mathbb{E}_{\Phi\sim \Psi_0}[G_{0}(\Psi_0)]] \\
&& = \sum_{i\in [k-1]}( \mathbb{E}_{\Psi_i}[\mathbb{E}_{\Phi\sim \Psi_i}[G_{i+1}(\Psi_{i+1})]]\\
&&\quad\quad- \mathbb{E}_{\Psi_i}[\mathbb{E}_{\Phi\sim \Psi_i}[G_{i}(\Psi_i)]]) \\
&& = \sum_{i\in [k-1]} \mathbb{E}_{\Psi_i}[\mathbb{E}_{\Phi\sim \Psi_i}[G_{i+1}(\Psi_{i+1})- G_{i}(\Psi_i)]] \\
&& \geq \sum_{i\in [k-1]} ((1-\epsilon)\frac{1}{k}(1-\frac{1}{k})^{k-(i+1)}g_{avg}(\pi^o) \\
&&\quad\quad-(1-\epsilon)\frac{1}{k}c_{avg}(\pi^o))\\
&& = \sum_{i\in [k-1]}((1-\epsilon)\frac{1}{k}(1-\frac{1}{k})^{k-(i+1)}g_{avg}(\pi^o))\\
&& \quad\quad- (1-\epsilon)c_{avg}(\pi^o)\\
&& \geq (1-\epsilon)(1-\frac{1}{e})g_{avg}(\pi^o) - (1-\epsilon)c_{avg}(\pi^o)\\
&& \geq (1-\frac{1}{e}-\epsilon)g_{avg}(\pi^o) - c_{avg}(\pi^o)
\end{eqnarray*} The first inequality is due to Lemma \ref{lem:b3}. $\Box$

\section{Non-monotone $g$:  Adaptive Random Distorted Greedy Policy}
  We next discuss the case when $g$ is non-negative adaptive submodular. We present an \emph{Adaptive Random Distorted Greedy Policy} $\pi^r$ for this case.
\subsection{Design of $\pi^r$} The detailed implementation of $\pi^r$ is listed in Algorithm \ref{alg:LPP2}. We first add a set $D$ of $k-1$ dummy items to the ground set, such that, for any $d \in D$, and any partial realization $\psi$, we have $\Delta(d \mid \psi) =0$. Let $E'=E\cup D$. $\pi^r$ runs round by round: Starting with an empty set. In each iteration $i\in[k-1]$, $\pi^{r}$  randomly selects an item from the set $M(\psi_i)$. Recall that $M(\psi_i)$ is a set of $k$ items that have the largest $H_i(\psi_i, \cdot)$, i.e.,
 \[M(\psi_i)\leftarrow \arg\max_{M\subseteq E'; |M|\leq k} \sum_{e\in E'}H_i(\psi_i, e)\]
After observing the state $\phi(e_i)$ of $e_i$, update  the current partial realization $\psi_{i+1}$ using  $\psi_{i}\cup\{\phi(e_i)\}$.  This process iterates until all  $k$ items have been selected.
\begin{algorithm}[hptb]
\caption{ Adaptive Random Distorted Greedy Policy $\pi^r$}
\label{alg:LPP2}
\begin{algorithmic}[1]
\STATE $S_0=\emptyset; i=0; \psi_0=\emptyset$.
\WHILE {$i < k$}
%\STATE observe $\psi_i$;
\STATE $M(\psi_i)\leftarrow \arg\max_{M\subseteq E'; |M|\leq k} \sum_{e\in E'}H_i(\psi_i, e)$;
\STATE sample $e_i$ uniformly at random from $M(\psi_i)$;
\STATE $S_i\leftarrow S_{i-1}\cup \{e_i\}$;
\STATE $\psi_{i+1}\leftarrow \psi_{i}\cup\{\phi(e_i)\}$;  $i\leftarrow i+1$;
\ENDWHILE
\RETURN $S_k$
\end{algorithmic}
\end{algorithm}

\subsection{Performance Analysis}
We first present three preparatory lemmas. The first lemma immediately follows from Lemma \ref{lem:a1}.
\begin{lemma}
\label{lem:c1}
In each iteration of $\pi^d$,
\begin{eqnarray*}
&&\mathbb{E}_{\Phi\sim \psi_i}[G_{i+1}(\Psi_{i+1})- G_{i}(\psi_i)] \\
&&= \mathbb{E}_{e_i}[H_i(\psi_i, e_i)]+\frac{1}{k}(1-\frac{1}{k})^{k-(i+1)} g(\psi_i)
\end{eqnarray*}
\end{lemma}

\begin{lemma}
\label{lem:c2}
In each iteration of $\pi^d$,
$
\mathbb{E}_{e_i}[H_i(\psi_i, e_i)] \geq \frac{1}{k}(1-\frac{1}{k})^{k-(i+1)} \mathbb{E}_{\Phi\sim \psi_i}[g_{avg}(\pi^o@\pi^d_i)-g_{avg}(\pi^d_i)]-\frac{1}{k}\mathbb{E}_{\Phi\sim \psi_i}[c_{avg}(\pi^o)]$.
\end{lemma}
\emph{Proof:}  Recall that $A_e$ is an indicator that $e$ is selected by the optimal solution $\pi^o$ conditioned on a partial realization $\psi_i$,
\begin{eqnarray*}
&& \mathbb{E}_{e_i}[H_i(\psi_i, e_i)]  \\
&& = \frac{1}{k}\sum_{e\in M(\psi_i)} ((1-\frac{1}{k})^{k-(i+1)} g(e \mid \psi_i)-c_{e})\\
&& \geq  \frac{1}{k}\sum_{e\in E'} \Pr[A_e=1] ((1-\frac{1}{k})^{k-(i+1)} g(e \mid \psi_i)-c_{e})\\
&& \geq \frac{1}{k}(1-\frac{1}{k})^{k-(i+1)}\mathbb{E}_{\Phi\sim \psi_i}[g_{avg}(\pi^o@\pi^d_i)-g_{avg}(\pi^d_i)]\\
&& \quad\quad-\frac{1}{k}\mathbb{E}_{\Phi\sim \psi_i}[c_{avg}(\pi^o)]
\end{eqnarray*} The equality is due to the design of $\pi^r$, i.e.,  it selects an item $e_i$ uniformly at random from $M(\psi_i)$. The first inequality is due to $\sum_{e\in E'}\Pr[A_e=1]\leq k$ since $\pi^o$ selects at most $k$ items, and $M(\psi_i)$ contains a set of $k$ items that have the largest $H_i(\psi_i, \cdot)$. The second inequality is due to $g$ is adaptive submodular and $c$ is modular. $\Box$

\begin{lemma}
\label{lem:c3}
In each iteration of $\pi^d$,
\begin{eqnarray*}
&&\mathbb{E}_{\Psi_i}[\mathbb{E}_{\Phi\sim \Psi_i}[G_{i+1}(\Psi_{i+1})- G_{i}(\Psi_i)]] \\
&&\geq \frac{1}{k}(1-\frac{1}{k})^{k-1}g_{avg}(\pi^o) -\frac{1}{k}c_{avg}(\pi^o)
\end{eqnarray*}
\end{lemma}
\emph{Proof:} We first show that for any partial realization $\psi_i$,
\begin{eqnarray}
&\mathbb{E}_{\Phi\sim \psi_i}[G_{i+1}(\Psi_{i+1})- G_{i}(\psi_i)] \label{eq:c1}\\
& \geq \frac{1}{k}(1-\frac{1}{k})^{k-(i+1)} \mathbb{E}_{\Phi\sim \psi_i}[g_{avg}(\pi^o@\pi^d_i)]-\frac{1}{k}\mathbb{E}_{\Phi\sim \psi_i}[c_{avg}(\pi^o)]~\nonumber
%&& \geq (\frac{1}{k}(1-\frac{1}{k})^{k-(i+1)} \mathbb{E}_{\Phi\sim \psi_i}[g_{avg}(\pi^o@\pi^d_i)]-\frac{1}{k}\mathbb{E}_{\Phi\sim \psi_i}[c_{avg}(\pi^o)])
\end{eqnarray}
Due to Lemma \ref{lem:c1}, we have
\begin{eqnarray*}
&&\mathbb{E}_{\Phi\sim \psi_i}[G_{i+1}(\Psi_{i+1})- G_{i}(\psi_i)] \\
&& = \mathbb{E}_{e_i}[H_i(\psi_i, e_i)]+\frac{1}{k}(1-\frac{1}{k})^{k-(i+1)} g(\psi_i)\\
&& \geq \frac{1}{k}(1-\frac{1}{k})^{k-(i+1)}\mathbb{E}_{\Phi\sim \psi_i}[g_{avg}(\pi^o@\pi^d_i)-g_{avg}(\pi^d_i)]\\
&& \quad\quad-\frac{1}{k}\mathbb{E}_{\Phi\sim \psi_i}[c_{avg}(\pi^o)]+\frac{1}{k}(1-\frac{1}{k})^{k-(i+1)} g(\psi_i)\\
&& = \frac{1}{k}(1-\frac{1}{k})^{k-(i+1)} \mathbb{E}_{\Phi\sim \psi_i}[g_{avg}(\pi^o@\pi^d_i)]\\
&& \quad\quad -\frac{1}{k}\mathbb{E}_{\Phi\sim \psi_i}[c_{avg}(\pi^o)] \\
%&& \geq (\frac{1}{k}(1-\frac{1}{k})^{k-(i+1)} \mathbb{E}_{\Phi\sim \psi_i}[g_{avg}(\pi^o@\pi^d_i)]-\frac{1}{k}\mathbb{E}_{\Phi\sim \psi_i}[c_{avg}(\pi^o)])
\end{eqnarray*}
The first inequality is due to Lemma \ref{lem:c2}. The second equality is due to $\mathbb{E}_{\Phi\sim \psi_i}[g_{avg}(\pi^d_i)] =  g(\psi_i)$. The last inequality is due to $g$ is non-negative. Now we are ready to prove this lemma.
\begin{eqnarray*}
&&\mathbb{E}_{\Psi_i}[\mathbb{E}_{\Phi\sim \Psi_i}[G_{i+1}(\Psi_{i+1})- G_{i}(\Psi_i)]] \\
&& \geq \mathbb{E}_{\Psi_i}[\frac{1}{k}(1-\frac{1}{k})^{k-(i+1)} \mathbb{E}_{\Phi\sim \Psi_i}[g_{avg}(\pi^o@\pi^d_i)]\\
&&\quad\quad-\frac{1}{k}\mathbb{E}_{\Phi\sim \Psi_i}[c_{avg}(\pi^o)]]\\
&& = \mathbb{E}_{\Psi_i}[\frac{1}{k}(1-\frac{1}{k})^{k-(i+1)} \mathbb{E}_{\Phi\sim \Psi_i}[g_{avg}(\pi^o@\pi^d_i)]] \\
&&\quad\quad-  \mathbb{E}_{\Psi_i}[\frac{1}{k}\mathbb{E}_{\Phi\sim \Psi_i}[c_{avg}(\pi^o)]]\\
&& = \frac{1}{k}(1-\frac{1}{k})^{k-(i+1)}g_{avg}(\pi^o@\pi^d_i) -\frac{1}{k}c_{avg}(\pi^o)\\
&& \geq \frac{1}{k}(1-\frac{1}{k})^{k-(i+1)}(1-\frac{1}{k})^{i}g_{avg}(\pi^o) -\frac{1}{k}c_{avg}(\pi^o)\\
&& = \frac{1}{k}(1-\frac{1}{k})^{k-1}g_{avg}(\pi^o) -\frac{1}{k}c_{avg}(\pi^o)
\end{eqnarray*}
The first inequality is due to (\ref{eq:c1}), and the second inequality is due to Lemma 1 in \cite{tang2021beyond}, where they show that $g_{avg}(\pi^o@\pi^d_i)\geq(1-\frac{1}{k})^{i}g_{avg}(\pi^o)$.  $\Box$

We next present the third main theorem of this paper.
\begin{theorem}
$g_{avg}(\pi^d) - c_{avg}(\pi^d) \geq (1-\frac{1}{e}-\epsilon)g_{avg}(\pi^o) - c_{avg}(\pi^o)$.
\end{theorem}
\emph{Proof:} According to the definition of $G_k$, we have $\mathbb{E}_{\Psi_k}[G_{k}(\Psi_k)] = \mathbb{E}_{\Psi_k}[(1-\frac{1}{k})^{0} g(\Psi_k)-c(\textrm{dom}(\Psi_k))]=g_{avg}(\pi^d) - c_{avg}(\pi^d)$ and $\mathbb{E}_{\Psi_0}[G_{0}(\Psi_0)] = \mathbb{E}_{\Psi_0}[(1-\frac{1}{k})^{k} g(\Phi(S_0))-c(\textrm{dom}(\Psi_0))]=0$. Hence,
\begin{eqnarray*}
&& g_{avg}(\pi^d) - c_{avg}(\pi^d) = \mathbb{E}_{\Psi_k}[G_{k}(\Psi_k)] - \mathbb{E}_{\Psi_0}[G_{0}(\Psi_0)]\\
&& = \mathbb{E}_{\Psi_{k-1}}[\mathbb{E}_{\Phi \sim \Psi_{k-1}}[G_{k}(\Psi_k)]]- \mathbb{E}_{\Psi_0}[\mathbb{E}_{\Phi\sim \Psi_0}[G_{0}(\Psi_0)]] \\
&& = \sum_{i\in [k-1]}( \mathbb{E}_{\Psi_i}[\mathbb{E}_{\Phi\sim \Psi_i}[G_{i+1}(\Psi_{i+1})]]\\
&&\quad\quad- \mathbb{E}_{\Psi_i}[\mathbb{E}_{\Phi\sim \Psi_i}[G_{i}(\Psi_i)]]) \\
&& = \sum_{i\in [k-1]} \mathbb{E}_{\Psi_i}[\mathbb{E}_{\Phi\sim \Psi_i}[G_{i+1}(\Psi_{i+1})- G_{i}(\Psi_i)]] \\
&& \geq \sum_{i\in [k-1]} (\frac{1}{k}(1-\frac{1}{k})^{k-1}g_{avg}(\pi^o) -\frac{1}{k}c_{avg}(\pi^o))\\
&& = \sum_{i\in [k-1]}\frac{1}{k}(1-\frac{1}{k})^{k-1}g_{avg}(\pi^o) - c_{avg}(\pi^o)\\
&& \geq \frac{1}{e}g_{avg}(\pi^o) - c_{avg}(\pi^o)
\end{eqnarray*} The first inequality is due to Lemma \ref{lem:c3}. $\Box$
\section{Conclusion}
In this paper, we study adaptive regularized submodular maximization problem. Because our objective function may take both negative and positive values, most existing technologies of submodular maximization do not apply to our setting. We develop a series of effective policies for our problem.
\bibliographystyle{siam}
\bibliography{reference}
\end{document}